%% file: paper.tex
\pdfoutput=1
\documentclass[11pt,letterpaper]{article}
\usepackage[margin=1in]{geometry}
\usepackage[T1]{fontenc}
\usepackage[utf8]{inputenc}
\usepackage{lmodern}
\usepackage{microtype}
\usepackage{amsmath,amssymb}
\usepackage{amsthm}
\usepackage{graphicx}
\usepackage{booktabs}
\usepackage{array}
\usepackage{longtable}
\usepackage{enumitem}
\usepackage[section]{placeins}
\usepackage[numbers,sort&compress]{natbib}
\usepackage[hidelinks]{hyperref}
\usepackage{tikz}
\usetikzlibrary{arrows.meta, positioning, fit, backgrounds, calc, shapes.geometric}

\definecolor{orxblue}{HTML}{0072B2}
\definecolor{orxorange}{HTML}{E69F00}
\definecolor{orxgreen}{HTML}{009E73}
\definecolor{orxred}{HTML}{D55E00}
\definecolor{orxpurple}{HTML}{CC79A7}
\tikzset{
  orx/.style={font=\sffamily\footnotesize, >={Stealth[round, length=4pt, width=3pt]}, line width=0.5pt,
    node distance=5mm and 7mm},
  block/.style={draw=black!55, fill=white, rounded corners=1pt, minimum height=6.5mm, inner sep=3pt, align=center},
  ours/.style={block, draw=orxblue, fill=orxblue!7, line width=0.7pt},
  muted/.style={block, draw=black!30, fill=black!3, text=black!65},
  flow/.style={->, draw=black!65},
  edgelbl/.style={font=\sffamily\scriptsize, text=black!60, inner sep=1.5pt, fill=white},
  stage/.style={draw=black!25, dashed, rounded corners=2pt, inner sep=4pt},
  stagelbl/.style={font=\sffamily\scriptsize\itshape, text=black!55},
}

\newcommand{\repourl}{https://github.com/chasemetoyer/found-by-search}
\newcommand{\repotag}{v1.1}
\newcommand{\repofile}[1]{\repourl/blob/\repotag/#1}
\newcommand{\src}[2]{Produced by \texttt{make #1}; data: \href{\repofile{#2}}{\nolinkurl{#2}}.}
\input{numbers.tex}
\makeatletter
\DeclareRobustCommand{\val}[1]{\@ifundefined{val@#1}{\PackageError{numbers}{Undefined value `#1'}{Run make numbers}}%
  {\href{\repofile{results/numbers.json}\#L\@nameuse{line@#1}}{\@nameuse{val@#1}}}}
\newcommand{\rawval}[1]{\@nameuse{val@#1}}
\makeatother
\newcommand{\pv}[1]{$p=$\,\val{#1.p}}
\newcommand{\ph}[1]{Holm $p=$\,\val{#1.ph}}
\newcommand{\REG}{\textnormal{\textsc{registered}}}
\newcommand{\PSD}{\textnormal{\textsc{pre-specified descriptive}}}
\newcommand{\EXP}{\textnormal{\textsc{exploratory/post hoc}}}
\newcommand{\DER}{\textnormal{\textsc{derived reference}}}
\newcommand{\HYP}{\textnormal{\textsc{post hoc hypothesis}}}

\newcommand{\paperTitle}{Found by Search, Broken by a Shortcut: Transfer and Tampering of Searched Intrinsic Rewards}
\title{\textbf{\paperTitle}}
\author{Chase J. Metoyer\\
Northeastern University\\
Boston, MA\\
\texttt{\href{mailto:metoyer.c@northeastern.edu}{metoyer.c@northeastern.edu}}\thanks{Code, data and the source of every
number: \url{https://github.com/chasemetoyer/found-by-search} (tag \texttt{v1.1}); archived at
\href{https://doi.org/10.5281/zenodo.23828842}{doi:10.5281/zenodo.23828842}. Each number in
this paper links to its line in \nolinkurl{results/numbers.json}; each table and figure names the command and file
that produce it.}}
\date{}

\begin{document}
\maketitle

\begin{abstract}
Can a search over innate drives find intrinsic rewards that help a reinforcement-learning agent learn, transfer to a
task it never saw, and resist an exploit that was absent during the search? Each candidate reward combines four
generic events (a new view, a pickup, a drop and a toggle), each with a weight, a relief and a refill rate. We ran
three pre-registered stages; the search protocol was locked in a public, timestamped commit before any search
result existed. CMA-ES and an equal-budget random search used \val{s2.runs} PPO training runs, and the final
evaluation and tamper test used \val{s3.runs}. On the held-out task, KeyCorridorS3R2, the CMA-ES-selected reward (E)
beat task-only PPO, episodic novelty and the random-search-selected reward (R) in all \val{kc.hits} registered
comparisons, all of them meaningful by the registered size rule. Among these generic rewards E transferred best;
the earlier paper's satiation reward, which on this task is given hand-wired key and locked-door events, did better
(\val{kc.S.solve.of} seeds solved against \val{kc.E.solve.of}; descriptive). On UnlockPickup, where that reward uses
generic events, E's mean area exceeded it by \val{up.EminusS.gap} (descriptive). On the training tasks there was no
registered difference between E and R: both E-over-R tests missed. A free reset action, absent during the search,
collapsed E's learning on DoorKey-8x8 (mean area \val{tam.off.area} without it, \val{tam.on.area} with it;
registered) to below task-only PPO (descriptive). The searched drives helped on every task tested, but a reward given
task-specific events did better on the held-out task, and selection gave E no protection against an exploit absent
from the search.
\end{abstract}


\section{Introduction}
\label{sec:introduction}

In our earlier paper \citep{metoyer2026satisfied}, PPO agents given a bounded, hand-designed homeostatic satiation reward
and a cheap action that reset their needs learned to press it and stopped solving the task, although solving was
optimal under that reward. That reward was designed by hand. The optimal-reward view suggests another route:
treat the reward as the outcome of a search over reward functions for whatever best serves the designer's goal
\citep{singh2010intrinsically}. This paper asks whether such a search avoids the failure, and what a searched reward
does on a task it was not selected on.

We search over innate drives. A candidate reward has four needs, tied to four generic events of the MiniGrid
interface \citep{chevalier2023minigrid}: seeing a new view, picking an object up, dropping it, and toggling a door or
box. Each need has a weight, a relief and a refill rate, and the reward is the signed reduction of a weighted
quadratic drive, the form the earlier paper used. We ask three questions.
\begin{itemize}[itemsep=1pt]
  \item \textbf{Help.} Do searched rewards help PPO \citep{schulman2017proximal} learn the tasks they were selected on?
  \item \textbf{Transfer.} Do they help on a task that was never used during the search?
  \item \textbf{Resistance.} Does a searched reward resist an exploit that was absent during the search?
\end{itemize}

\begin{figure}[t]
\centering
\begin{tikzpicture}[orx]
  \node[muted] (s1) {Stage~1\\calibration,\\detector, tasks};
  \node[block, right=5mm of s1, yshift=6mm] (cma) {CMA-ES\\\val{s2.gens} generations $\times$ \val{s2.pop}};
  \node[block, below=3mm of cma] (rand) {Random search\\\val{s2.cands} uniform draws};
  \node[block, right=5mm of $(cma.east)!0.5!(rand.east)$, anchor=west] (fin) {Finalists\\top \val{s2.finalists} per method,\\\val{s2.reevalseeds} new seeds per task};
  \node[ours, right=5mm of fin] (frz) {Frozen\\E (cand.\ \val{win.E})\\R (cand.\ \val{win.R})\\$P_0$ recorded};
  \node[block, right=5mm of frz, yshift=7mm] (ev) {Final evaluation\\E, R, T, N; S descriptive\\\val{d.ntasks} tasks $\times$ \val{s3.seeds} seeds};
  \node[block, below=3mm of ev] (tm) {Tamper retraining\\E, reset on vs off\\DoorKey-8x8, \val{s3.seeds} seeds};
  \draw[flow] (s1.east) -- ++(3mm,0) |- (cma.west);
  \draw[flow] (s1.east) -- ++(3mm,0) |- (rand.west);
  \draw[flow] (cma.east) -- (fin.west |- cma.east);
  \draw[flow] (rand.east) -- (fin.west |- rand.east);
  \draw[flow] (fin) -- (frz);
  \draw[flow] (frz.east) -- ++(3mm,0) |- (ev.west);
  \draw[flow] (frz.east) -- ++(3mm,0) |- (tm.west);
  \begin{scope}[on background layer]
    \node[stage, fit=(cma)(rand)(fin)(frz)] (st2) {};
    \node[stage, fit=(ev)(tm)] (st3) {};
  \end{scope}
  \node[stagelbl, above=0.5mm of st2] {Stage~2: search on DoorKey-8x8 and UnlockPickup};
  \node[stagelbl, above=0.5mm of st3] {Stage~3};
  \node[stagelbl, below=0.5mm of st2, align=center] {protocol (Stages~2--3) locked and pushed publicly before any search result};
\end{tikzpicture}
\caption{\textbf{Three pre-registered stages.} Stage~1 fixed the detector and selected tasks. In Stage~2, every
candidate was trained on both training tasks with \val{s2.searchseeds} seeds each; the top \val{s2.finalists}
candidates per method were re-evaluated on fresh seeds and the best re-evaluated candidate per method was frozen
as E or R. The Stage~2 lock also registered every Stage~3 test. Stage~3 evaluated E, R, task-only PPO (T) and
episodic novelty (N) on the two training tasks and on the held-out KeyCorridorS3R2, with the earlier paper's satiation
reward (S) as a descriptive arm, and retrained E with a free reset action. Seed sets never overlap across roles.}
\label{fig:design}
\end{figure}
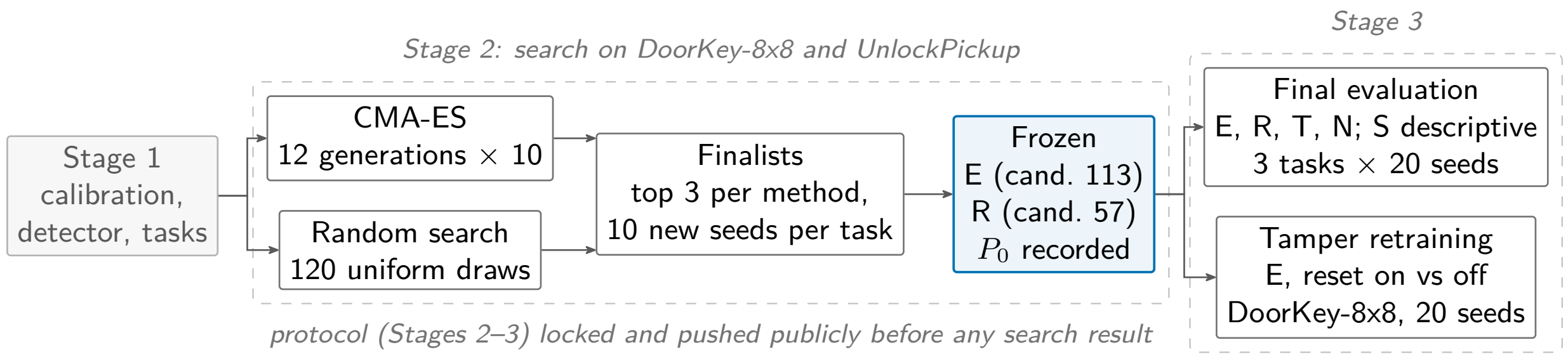

We answered them in three pre-registered stages (Figure~\ref{fig:design}). Stage~1 calibrated tasks and fixed a
corrected event detector. Stage~2 ran CMA-ES \citep{hansen2001completely} and an equal-budget random search over the
twelve drive parameters and selected one reward from each. Stage~3 compared the CMA-ES-selected reward (E) and the
random-search-selected reward (R) with task-only PPO (T) and episodic novelty (N) on two training tasks and one
held-out task, retrained E with a free reset action, and ran the earlier paper's satiation reward (S) as a descriptive
reference. Comparisons of E and R compare two selected genomes; one draw from each search method cannot rank the
methods.

Our contributions, each with its evidence level:
\begin{itemize}[itemsep=1pt]
  \item \textbf{Transfer to one held-out task, bounded by a hand-wired reference (Section~\ref{sec:transfer}).} On
  KeyCorridorS3R2, E beat T, N and R in all \val{kc.hits} registered comparisons, each meaningful by the
  registered size rule (\REG). S, whose KeyCorridor events are hand-wired (key pickups only, and only the opening of
  the initially locked door) and whose door need refills, solved \val{kc.S.solve} seeds against E's
  \val{kc.E.solve}; on UnlockPickup, where S uses generic events, E's mean area was higher (\PSD).
  \item \textbf{No registered difference between the selected rewards on the training tasks
  (Sections~\ref{sec:search} and~\ref{sec:training}).} Both E-over-R tests missed (\REG). Every CMA-ES finalist
  fell on re-evaluation and every random-search finalist rose (\EXP).
  \item \textbf{No protection against an exploit absent during selection (Section~\ref{sec:tamper}).} A free
  reset action collapsed E's learning (\REG), to below task-only PPO (\PSD).
  \item \textbf{What the selected rewards encode (Section~\ref{sec:genomes}).} E pays the first pickup, drop and toggle of
  each episode and refills only its view need (a record of the frozen genome). Across all searched candidates, positive weights on pickup and toggle
  were associated with higher fitness across all candidates and among random-search candidates alone; the drop
  association reverses among random-search candidates (\EXP).
  \item \textbf{A public pre-registration lock (Section~\ref{sec:prereg}, Appendix~\ref{app:timeline}).} The earlier paper's
  lock times were self-reported. Here the Stage~2 protocol, which also registered every Stage~3 test, was committed
  and pushed to a public repository before any search result existed (pushed \val{reg.s2.push}).
  \item \textbf{Every registered test and rule is reported, including misses (Section~\ref{sec:wrong},
  Appendix~\ref{app:outcomes}).}
\end{itemize}

Results carry one of three labels. \REG{} marks confirmatory tests and hit rules fixed in the locked protocol.
\PSD{} marks descriptive quantities named in the locked protocol or in amendments committed before the data
existed. \EXP{} marks everything else. We also label arithmetic references (\DER) and untested explanations (\HYP).
Event definitions differ from the earlier paper's archived detector, so comparisons with the earlier paper are qualitative only.

\section{Background and related work}
\label{sec:related}

\paragraph{Optimal, searched and learned rewards.} \citet{singh2010intrinsically} define an optimal-reward framework
in which primary reward functions are selected for evolutionary success across environments, and conclude that
intrinsic and extrinsic reward differ along a continuum rather than in kind. \citet{niekum2010genetic} search
reward functions with genetic programming and name distributions of environments, nonstationary environments and
short agent lifetimes as settings where reward search should help. \citet{sorg2010reward} solve the optimal-reward
problem online by gradient ascent for agents with limitations. \citet{zheng2018learning} learn intrinsic-reward
parameters for policy-gradient agents by gradient ascent on the extrinsic return, and \citet{xu2018meta} adapt
return parameters such as the discount online with meta-gradients. Several studies test generalization directly.
\citet{zheng2020what} learn intrinsic rewards over many lifetimes and report that the learned rewards generalize to
other kinds of agents and to changes in the environment's dynamics. \citet{alet2020meta} meta-learn curiosity
algorithms as programs and report two that perform on par with or better than published curiosity algorithms in
disparate domains. \citet{oh2020discovering} discover an update rule on toy environments that generalizes to Atari
games, and \citet{oh2025discovering} discover, from a population of agents across many environments, a rule that
surpasses existing rules on Atari and outperforms several algorithms on benchmarks not seen during discovery.

\paragraph{Language-model reward design (contrast).} \citet{ma2024eureka} use large language models to write reward
code and improve it by evolutionary search, and report rewards that outperform expert-written ones on most of their
tasks. Our search space is different in kind: twelve numbers in a fixed homeostatic form, with no language model.

\paragraph{Intrinsic motivation and homeostasis.} Exploration bonuses include pseudo-counts
\citep{bellemare2016unifying}, prediction error at scale \citep{burda2019largescale}, random network distillation
\citep{burda2019rnd} and impact-driven bonuses evaluated in procedurally generated MiniGrid tasks
\citep{raileanu2020ride}. \citet{burda2019largescale} also gave a curiosity agent a television whose channel it
could change; the television drastically slowed learning, although agents sometimes still converged to the
extrinsic reward. \citet{keramati2014homeostatic} define primary reward as the fulfilment of physiological needs and prove
that seeking such reward is equivalent to seeking physiological stability. Our predecessor study tested a bounded satiation reward of
this kind and found it wireheaded through a reset action \citep{metoyer2026satisfied}.

\paragraph{Reward hacking and tampering.} \citet{amodei2016concrete} list avoiding reward hacking among concrete
accident-risk problems, and \citet{skalse2022defining} give a formal definition of reward hacking. \citet{everitt2021reward}
study when an agent has an instrumental goal to tamper with its reward process and give design principles against
reward-function tampering and against tampering with the reward function's input. Our reset action changes the
need state the reward reads, so in their terms it is input tampering, as in the earlier paper.

\paragraph{Random search and CMA-ES.} \citet{bergstra2012random} show that random trials are more efficient than a
grid when only a few hyperparameters matter, and propose random search as the natural baseline for adaptive
methods. CMA-ES adapts a full covariance matrix of a Gaussian search distribution
\citep{hansen2001completely,hansen2016cma}; we use the pycma implementation \citep{hansen2026pycma}.

\paragraph{How this paper differs.} The generalization studies above learn reward networks, programs or update rules
and test them on new agents, dynamics or benchmarks. We search a small, interpretable space of drives, compare the
CMA-ES-selected reward with an equal-budget random-search selection, test transfer to one held-out task and then test
the same reward against an exploit absent from its selection environment. We are not aware of a prior study that
pairs a held-out transfer test and a tamper test of the same searched reward under a public pre-registration.

\section{Setup}
\label{sec:setup}

\subsection{The reward}
\label{sec:reward}
A candidate reward has needs $h=(h_\text{view},h_\text{pickup},h_\text{drop},h_\text{toggle})\in[0,1]^4$, each
starting at $h_0=\val{d.h0}$ in every episode. At each step the needs grow by the refill rates $g$ and are relieved
by the events $e_t\in\{0,1\}^4$ in proportion to the reliefs $s$, with a single clip, and the reward is the drop
in a signed, weighted quadratic drive:
\begin{equation}
h_{t+1}=\operatorname{clip}\!\left(h_t+g-s\odot e_t,\,0,\,1\right),\qquad
V(h)=\sum_k w_k h_k^2,\qquad r^\text{int}_t=V(h_t)-V(h_{t+1}).
\label{eq:reward}
\end{equation}
PPO trains on the task reward plus $\val{d.coef}\times r^\text{int}_t$. Weights $w_k$ lie in $[-1,1]$, reliefs in
$[0,1]$, and each refill rate is either zero or in $[\val{d.gmin},\val{d.gmax}]$ per step. Summed over an episode the
reward telescopes (Stage~1 derivation, re-derived in Appendix~\ref{app:hyper}): for any policy and any
$\gamma\in(0,1]$, the intrinsic return lies between $V(h_0)-\sum_k\max(w_k,0)$ and $V(h_0)-\sum_k\min(w_k,0)$. When
every weight is positive, as in E, the upper bound is $V(h_0)$, the drive at the start. This is also the payoff of
a reset at the first step, $P_0=\sum_k w_k h_0^2$, which the tamper prediction used.

\subsection{Events and the corrected detector}
\label{sec:detector}
The four events are generic. A \emph{new view} is a $7\times7$ egocentric observation not yet seen in the episode; a
\emph{pickup} is any change from carrying nothing to carrying an object; a \emph{drop} is an executed drop; a
\emph{toggle} is any door changing its open or locked state, including closing and reopening, or a box being opened.
Stage~1 fixed four detector choices: events on an episode's final step count; box openings count as toggles; a
pickup that itself completes the task is excluded from the paid pickup event; and the detector attaches only after
the initial reset. The earlier paper's archived detector differed in ways documented by an audit of its code committed with the
Stage~1 materials: it skipped pickup and door events on an episode's final step, its door cache was empty during each
parallel environment's first training episode, it did not detect box openings, and it had no drop event. Its event
counts and ours are therefore not interchangeable.

\subsection{Learner and tasks}
\label{sec:tasks}
The learner is the earlier paper's PPO \citep{schulman2017proximal}, in the single-file style of CleanRL
\citep{huang2022cleanrl}, with the earlier paper's settings (Appendix~\ref{app:hyper}): \val{ppo.envs} parallel
environments, per-minibatch advantage normalization, and nominal 2M environment steps per run
(\val{d.actual} actual steps, in whole rollouts). Tasks are MiniGrid environments \citep{chevalier2023minigrid}. Stage~1
applied a fixed selection rule to task-only and episodic-novelty baselines: a training task was kept if its better
baseline mean area at 1M steps was at most \val{s1.easy} and some baseline seed had positive task success by 2M; a
held-out task was kept if its better baseline area at 2M lay between \val{s1.hard} and \val{s1.easy} (Section~\ref{sec:calibration}).

\subsection{Search}
\label{sec:searchsetup}
Both methods search a shared latent box $z\in[0,1]^{12}$, decoded identically: $w=2z-1$, $s=z$, and with $u=2z-1$ the
refill is zero if $u<0$ and $\val{d.gmin}\cdot(\val{d.gmax}\,/\,\val{d.gmin})^{u}$ otherwise, so a uniform draw has
zero refill on each event with probability one half. CMA-ES \citep{hansen2001completely} ran in pycma
\citep{hansen2026pycma} with population \val{s2.pop} for \val{s2.gens} generations, initial mean \val{s2.mean0} in
every coordinate, initial step size \val{s2.sigma}, pycma's default bound handling, no restarts and optimizer seed
\val{s2.cmaseed}. A generation was requested only after every run of the previous one was recorded, and any resume
had to reproduce every recorded candidate exactly. Random search drew \val{s2.cands} candidates uniformly from the
same box (generator seed \val{s2.randseed}), fixed and committed before any search evidence, and evaluated them in
batches of \val{s2.pop}.

A candidate's \emph{fitness} is its mean training-curve area over the \val{d.ntrain} training tasks with
\val{s2.searchseeds} seeds each, equally weighted. A run's area is the arithmetic mean of its \val{d.rows} logged
training-success values; each value is the success rate over the last \val{d.window} completed training episodes, and a
missing early value counts as zero. All candidates of one CMA-ES generation, or one random-search batch, shared
a seed panel (seeds \val{seeds.cma} and \val{seeds.rand}). The top \val{s2.finalists} candidates per method by
fitness were re-evaluated on \val{s2.reevalseeds} fresh seeds per training task (seeds \val{seeds.fincma} and
\val{seeds.finrand}), and the finalist with the highest re-evaluated mean area was frozen as that method's reward. In
all, Stage~2 trained \val{s2.runs} runs: \val{s2.search} search runs and \val{s2.reeval} re-evaluation runs.

\subsection{Final evaluation}
\label{sec:evalsetup}
Stage~3 trained five arms on DoorKey-8x8, UnlockPickup and the held-out KeyCorridorS3R2, with \val{s3.seeds} fresh
seeds per arm and task (\val{seeds.final}): E, R, task-only PPO (T) and episodic novelty with coefficient
\val{d.Ncoef} (N), plus the earlier paper's satiation reward with growth \val{d.Sgrowth} and the corrected detector (S). S
was registered as descriptive only. Its three needs (explore, grasp, open) differ from E's and R's in two ways that
bear on the held-out comparison. On KeyCorridor tasks its grasp event counts only key pickups and its open event counts
only the opening of a door that was locked at the start of the episode; on DoorKey-8x8 and UnlockPickup it uses the
generic pickup and toggle events. And all of its needs refill.

The registered endpoint is the training-curve area. Each comparison is a one-sided Mann--Whitney $U$ test of E
against a comparator, exact when the pooled areas have no ties and otherwise an independent-label permutation of
$U$ with \val{d.resamples} resamples and seed 0. A permutation $p$ of \val{d.pfloor} is the smallest attainable with
that many resamples. The primary family is E against T, N and R on KeyCorridorS3R2 (Holm, $m=\val{d.holm1}$); the
secondary family is the same comparisons on the two training tasks (Holm, $m=\val{d.holm2}$). A comparison is a
\emph{hit} when its Holm-adjusted $p$ is below \val{d.alpha}, and a \emph{meaningful improvement} when it is a hit and the observed mean
area of E exceeds the comparator's by at least \val{d.meaningful}. Solve counts (mean training success of at least
\val{d.solvethr} over the final \val{d.finalrows} logged values) carry Clopper--Pearson 95\% intervals and one-sided
Fisher exact $p$-values, which the protocol calls exploratory and which supply no hits. Median first-solved steps
are censored at the nominal budget. Runs used whatever CPU type the provider assigned, and same-seed results were
not bit-identical across CPU types, so every comparison is unpaired.

\subsection{Tamper test}
\label{sec:tampersetup}
E was retrained from scratch on DoorKey-8x8 with \val{s3.seeds} seeds per condition (\val{seeds.tamper}), with and
without a free reset action. With the action enabled, MiniGrid's no-op ``done'' action sets every need to zero before
the reward is computed, so pressing it pays the whole current drive. Before any tamper data, the protocol fixed the
prediction by the sign of $P_0=\sum_k w_k h_0^2$, computed from E's frozen float32 weights: $P_0=\val{tam.P0full}>0$,
so the registered prediction was \emph{collapse}, tested as a one-sided Mann--Whitney test that the mean area without
the action exceeds the mean area with it, with no fallback if it missed. A descriptive amendment, public before the
Stage~3 launch, added press counts per episode and per step, intrinsic return per episode against $P_0$, policy
entropy, a comparison with task-only PPO on DoorKey-8x8 and an ordered label: \emph{wireheading-like} if the tamper-on
mean area is at least \val{lab.gap} below task-only PPO's and the across-seed median presses per completed episode
exceed \val{lab.press} in every logging window of the final 10\% of training. These are descriptive and supply no
hits.

\subsection{Pre-registration}
\label{sec:prereg}
Stage~1 was locked at \val{reg.s1.lock} (pushed publicly \val{reg.s1.push}). The Stage~2 protocol was locked at
\val{reg.s2.lock} in commit \val{reg.s2.commit} of a public GitHub repository, pushed \val{reg.s2.push}, before any
search run; it froze the code, the random-search candidate list, the search settings, the winner rule and every
Stage~3 test, family and seed. Later changes went into timestamped amendments, none of which changed a scientific
choice (Appendix~\ref{app:timeline}). The Stage~3 descriptive amendment was carried by public commit
\val{reg.r1.commit} (verified on GitHub at \val{reg.r1.verified}) and unchanged in \val{reg.r2.commit} (verified at
\val{reg.r2.verified}); the Stage~3 science run started at \val{reg.launch} on the same day.

\section{Results}
\label{sec:results}

\subsection{Calibration and task selection}
\label{sec:calibration}

\begin{table}[t]
\centering
\small
\caption{\textbf{Stage~1 task selection (\REG{} rule, applied mechanically).} Best baseline mean training-curve
area over task-only PPO and episodic novelty, \val{s1.seeds} calibration seeds per arm, at the budget the rule uses. A training
candidate also needed positive task success by 2M in some baseline seed; UnlockPickup met that condition. \src{table1}{results/table1_calibration.csv}}
\label{tab:calibration}
\footnotesize
\input{tables/calibration.tex}
\end{table}

\begin{itemize}[itemsep=1pt]
  \item \textbf{Kept} (\REG, Table~\ref{tab:calibration}): DoorKey-8x8 and UnlockPickup as training tasks, and
  KeyCorridorS3R2 as the held-out task. \val{s1.rejected} candidates were rejected, \val{s1.tooeasy} as too easy and \val{s1.toohard} as too hard.
  \item UnlockPickup's baselines barely learned (best area \val{s1.UnlockPickup.area}); the training rule had no lower
  bound, and we chose equal task weighting in fitness.
  \item \textbf{Baseline separation at 1M} (\PSD; a registered two-sided test with no prediction): episodic novelty
  did not separate from task-only PPO on any training candidate (Appendix~\ref{app:outcomes}); we then set the
  fitness budget to 2M.
\end{itemize}

\subsection{Search}
\label{sec:search}

\begin{figure}[t]
\centering
\includegraphics[width=5.5in]{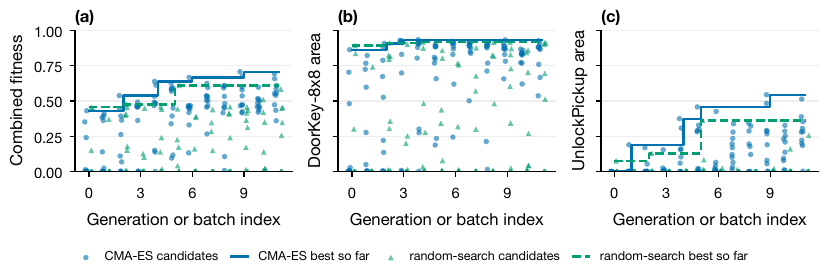}
\caption{\textbf{Both searches found candidates that learned DoorKey-8x8; CMA-ES reached higher search scores on
UnlockPickup (\val{s2.searchseeds}-seed search scores; not confirmed on re-evaluation, Table~\ref{tab:finalists}).} Search fitness of all \val{s2.candtotal} candidates (\val{s2.searchseeds} seeds per task each) by
CMA-ES generation or random-search batch, with each method's best score so far. Candidates in one generation or
batch shared a seed panel; the random-search batch index has no optimization meaning. Search scores are the
selection signal, not independent estimates (\PSD{} data, \EXP{} display). \src{figure2}{results/figure2_search.csv}}
\label{fig:search}
\end{figure}

\begin{table}[t]
\centering
\small
\caption{\textbf{Every CMA-ES finalist fell on re-evaluation; every random-search finalist rose.} Search fitness
(\val{s2.searchseeds} seeds per training task) and re-evaluated fitness (\val{s2.reevalseeds} fresh seeds per training
task) of the \val{s2.finalists} finalists per method; the re-evaluated score selects the winner, so it is also a
selection-panel score. Values \PSD; the change column is an \EXP{} display. \src{table2}{results/table2_finalists.csv}}
\label{tab:finalists}
\footnotesize
\input{tables/finalists.tex}
\end{table}

\begin{figure}[t]
\centering
\includegraphics[width=3.25in]{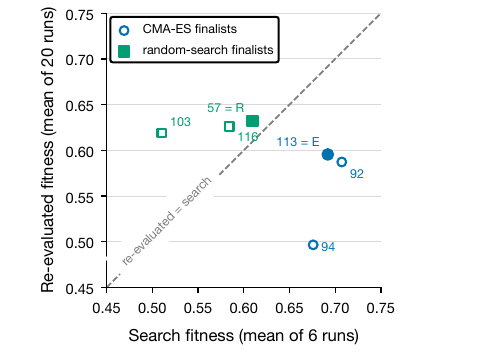}
\caption{\textbf{Search scores overstated the CMA-ES finalists.} Finalist search fitness against re-evaluated
fitness; filled markers are the frozen winners E and R. Points below the dashed line scored lower on fresh seeds.
\PSD{} values; the winner's-curse reading in the text is \EXP. \src{figure3}{results/figure3_finalists.csv}}
\label{fig:finalists}
\end{figure}

\begin{itemize}[itemsep=1pt]
  \item \textbf{Search scores} (\PSD, Figure~\ref{fig:search}). The best CMA-ES candidate (\val{cand.cma.bestid})
  scored \val{cand.cma.best} and the best random-search candidate (\val{cand.random.bestid}) scored
  \val{cand.random.best}.
  \item \textbf{Re-evaluation} (\PSD{} values, Table~\ref{tab:finalists}, Figure~\ref{fig:finalists}). All
  \val{fin.cmafell} CMA-ES finalists scored lower on fresh seeds (changes \val{fin.cma92.change},
  \val{fin.cma113.change} and \val{fin.cma94.change}), and all \val{fin.randrose} random-search finalists scored
  higher (\val{fin.random57.change}, \val{fin.random116.change} and \val{fin.random103.change}).
  \item \textbf{Winners} (\REG{} rule). E is CMA-ES candidate \val{win.E} (re-evaluated \val{fin.cma113.reeval}) and R
  is random-search candidate \val{win.R} (re-evaluated \val{fin.random57.reeval}). On their respective re-evaluation
  panels, which used different seeds (\val{seeds.fincma} and \val{seeds.finrand}), E minus R was \val{fin.EminusR}
  (\PSD); this is not an independent comparison.
  \item \textbf{Reading} (\EXP). The CMA-ES drop fits a winner's curse from selecting on noisy
  \val{s2.searchseeds}-seed estimates, but that alone does not predict the random-search rises; the different
  re-evaluation seed panels are a competing explanation, and CPU assignment adds noise. We tested neither.
\end{itemize}

\subsection{Transfer to the held-out task (registered primary family)}
\label{sec:transfer}

\begin{table}[t]
\centering
\small
\caption{\textbf{Final evaluation, \val{s3.seeds} seeds per arm and task (\PSD{} descriptives; the registered tests
are in Table~\ref{tab:comparisons}).} Mean training-curve area with its standard deviation across seeds, solved
seeds with Clopper--Pearson 95\% intervals, and median first-solved step, censored at the nominal budget.
$^\dagger$S is descriptive by design. On KeyCorridor tasks its grasp event counts only key pickups and its open event
counts only the opening of the door locked at the start of the episode, and its door need refills; on DoorKey-8x8
and UnlockPickup it uses the generic pickup and toggle events, and on UnlockPickup E's mean area exceeded S's. \src{table3}{results/table3_final.csv}}
\label{tab:final}
\input{tables/final.tex}
\end{table}

\begin{table}[t]
\centering
\footnotesize
\setlength{\tabcolsep}{4pt}
\caption{\textbf{Every planned comparison.} One-sided Mann--Whitney tests of E against each comparator on mean
training-curve area (\REG); Holm within each family. ``Meaningful'' additionally requires an observed mean gap of
at least \val{d.meaningful}. A permutation $p$ of \val{d.pfloor} is the floor for \val{d.resamples} resamples. The
last column is the protocol's exploratory one-sided Fisher test on solve counts (\PSD), which supplies no hits. \src{table4}{results/table4_comparisons.csv}}
\label{tab:comparisons}
\input{tables/comparisons.tex}
\end{table}

\begin{figure}[t]
\centering
\includegraphics[width=5.5in]{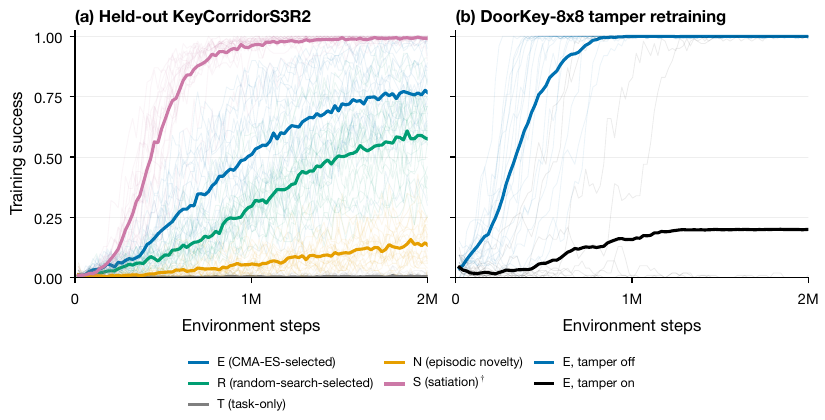}
\caption{\textbf{E learned the held-out task better than T, N and R, and S, whose KeyCorridor events are hand-wired,
learned it best; a free reset action collapsed E's learning on DoorKey-8x8.} Training success (rolling success over
the last \val{d.window} completed training episodes) for every seed (thin) and the seed mean (thick), \val{s3.seeds} seeds per
line. (a) KeyCorridorS3R2, never used during the search. (b) E retrained with and without the reset action. The
registered endpoints are the curve areas (Table~\ref{tab:comparisons}); on UnlockPickup, where S uses generic events,
E's mean area exceeded S's (Table~\ref{tab:final}). $^\dagger$On KeyCorridor tasks S's events are hand-wired (key pickups only; only the opening of the initially locked
door) and its door need refills. The registered Stage~3 figure redrawn from the same data, with relabelled legend
entries, a new tamper-on colour, a fuller panel (b) label and a shorter height. \src{figure4}{results/figure4_heldout_tamper.csv}}
\label{fig:curves}
\end{figure}

On KeyCorridorS3R2, E beat every registered comparator (\REG, Tables~\ref{tab:final} and~\ref{tab:comparisons},
Figure~\ref{fig:curves}a):
\begin{itemize}[itemsep=1pt]
  \item E vs T: mean gap \val{kc.ET.diff} (\pv{kc.ET}; \ph{kc.ET}); hit, meaningful.
  \item E vs N: mean gap \val{kc.EN.diff} (\pv{kc.EN}; \ph{kc.EN}); hit, meaningful.
  \item E vs R: mean gap \val{kc.ER.diff} (\pv{kc.ER}; \ph{kc.ER}); hit, meaningful.
  \item E solved \val{kc.E.solve} seeds; R, T and N solved \val{kc.R.solve}, \val{kc.T.solve} and \val{kc.N.solve} (\PSD).
\end{itemize}
The $p$-values of \val{kc.ET.p} are at the permutation floor, so they bound rather than estimate the evidence.
Among the generic-event arms, E transferred best (\REG{} comparisons). The earlier paper's satiation reward did better than E on this task: S
solved \val{kc.S.solve} seeds with mean area \val{kc.S.area}, against E's \val{kc.E.solve} and \val{kc.E.area} (\PSD;
S was not tested against E). S differs from E on KeyCorridor in two ways at once: its events are hand-wired (a key
pickup and the opening of the initially locked door are its only grasp and open events), and its door need refills.
On UnlockPickup, where S uses generic events, E's mean area exceeded S's by \val{up.EminusS.gap} (\PSD{}). The S result
bounds the transfer claim: a searched reward over generic events transferred better than the generic baselines, but
not as well as a reward told which events matter on this task. The transfer claim rests on this single held-out task.

\begin{figure}[t]
\centering
\includegraphics[width=5.5in]{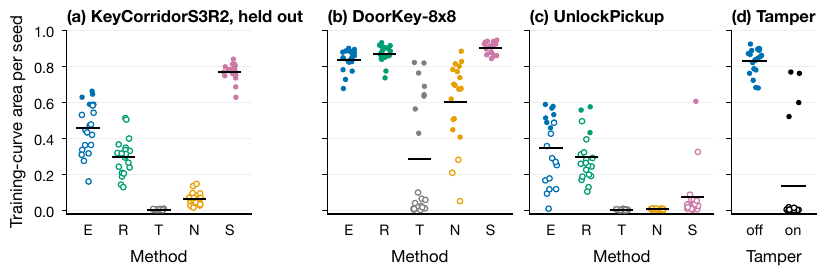}
\caption{\textbf{Per-seed areas show the same ordering as the means, with wide spread for E on the held-out task.}
Training-curve area of every seed; filled markers are solved seeds, hollow markers unsolved; bars are seed means,
\val{s3.seeds} seeds per arm. (a)--(c) Final evaluation; S on KeyCorridorS3R2 has hand-wired events, while on UnlockPickup, where S uses generic
events, E's areas were higher. (d) E retrained
on DoorKey-8x8 with the reset action off and on. \PSD. \src{figure5}{results/figure5_per_seed.csv}}
\label{fig:perseed}
\end{figure}

\subsection{Training tasks (registered secondary family)}
\label{sec:training}
\val{sec.hits} of the \val{d.holm2} secondary comparisons hit (\REG, Table~\ref{tab:comparisons}). \textbf{Both E-over-R comparisons
missed}:
\begin{itemize}[itemsep=1pt]
  \item \textbf{DoorKey-8x8, E vs R: miss.} Mean gap \val{dk.ER.diff} (\pv{dk.ER}; \ph{dk.ER}). R's mean area was the
  higher.
  \item \textbf{UnlockPickup, E vs R: miss.} Mean gap \val{up.ER.diff} (\pv{up.ER}; \ph{up.ER}). The observed gap
  passed the \val{d.meaningful} size rule but the rank test did not, so the comparison is neither a hit nor a meaningful
  improvement. A miss does not establish that E and R are equivalent.
  \item E beat T and N on both training tasks: DoorKey gaps \val{dk.ET.diff} and \val{dk.EN.diff} (\ph{dk.ET};
  \ph{dk.EN}); UnlockPickup gaps \val{up.ET.diff} and \val{up.EN.diff} (\ph{up.ET}; \ph{up.EN}).
  \item S (\PSD): on DoorKey-8x8 S's mean area (\val{dk.S.area}) exceeded E's (\val{dk.E.area}); on UnlockPickup,
  where S uses generic events, it was far below (\val{up.S.area} against \val{up.E.area}).
\end{itemize}

\subsection{Tamper test (registered, with pre-specified descriptives)}
\label{sec:tamper}

\begin{table}[t]
\centering
\small
\caption{\textbf{A free reset action collapsed E's learning, and the tamper-on agents collected nearly the whole
episode ceiling.} E retrained on DoorKey-8x8, \val{s3.seeds} seeds per condition. Press rates and returns are means
over seeds. The final 10\% of training is the last tenth of environment steps. Level: \REG{} or \PSD. \src{table5}{results/table5_tamper.csv}}
\label{tab:tamper}
\input{tables/tamper.tex}
\end{table}

\begin{figure}[t]
\centering
\includegraphics[width=5.5in]{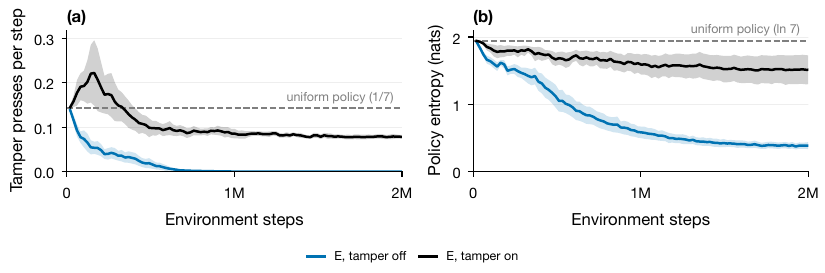}
\caption{\textbf{Tamper-on agents kept pressing and kept high policy entropy; tamper-off agents stopped pressing
and became decisive.} (a) Share of environment steps on the reset action and (b) PPO policy entropy over training,
mean over \val{s3.seeds} seeds per condition (all seeds in every logging window) with 95\% intervals. Dashed lines
are arithmetic references for a uniform policy over the 7 actions (\DER), not measured policies. Without the tamper
flag the action is a no-op, but its presses are still counted. \PSD. \src{figure6}{results/figure6_tamper_dynamics.csv}}
\label{fig:tamperdyn}
\end{figure}

\begin{itemize}[itemsep=1pt]
  \item \textbf{Collapse} (\REG). $P_0=\val{tam.P0}>0$ fixed the collapse branch before any tamper data. Mean area
  fell from \val{tam.off.area} without the reset action to \val{tam.on.area} with it (difference \val{tam.diff};
  $U=\val{tam.U}$, $p=$\,\val{tam.p}, the permutation floor). Hit. Seeds solved: \val{tam.solve.off} against
  \val{tam.solve.on} (\PSD).
  \item \textbf{Worse than task-only PPO} (\PSD). Tamper-on E's mean area was \val{tam.vsT.gap} below task-only PPO's
  on DoorKey-8x8 (\val{dk.T.area}). The two panels used different seeds (\val{seeds.tamper} and \val{seeds.final}) and
  no test was registered. This is qualitatively consistent with the earlier paper's finding that a self-fillable
  satisfaction signal was worse than none; detectors, rewards and seeds differ.
  \item \textbf{Label} (\PSD). The pre-specified label is \emph{\val{tam.label}}: the area rule held, and the median
  presses per completed episode exceeded \val{lab.press} in all \val{tam.windows} final-window logs (window medians
  \val{tam.medmin} to \val{tam.medmax}).
  \item \textbf{The press criterion is weak} (\PSD{} numbers, our reading). More than one press per episode is easy to
  exceed in long, failing episodes, as the earlier paper found for its own labels. In the final 10\% of training,
  tamper-on agents pressed on \val{tam.on.pps10} of steps, below the \val{tam.uniform} a uniform policy would press
  (\DER). Early in training the rate rose above the uniform rate (Figure~\ref{fig:tamperdyn}a). The area gap, not the
  label, is the primary evidence.
  \item \textbf{Intrinsic return against the ceiling} (\PSD). Mean intrinsic return per completed episode was
  \val{tam.on.iret} with the reset action and \val{tam.off.iret} without it, against $P_0=\val{tam.P0five}$, the
  ceiling for E's all-positive weights. No tamper-on episode exceeded $P_0$. One tamper-off episode exceeded it by
  \val{tam.off.excess}, a floating-point rounding margin; no episode in either condition exceeded 0.64.
  \item \textbf{Entropy} (\PSD). Mean policy entropy over training was \val{tam.on.ent} nats with the reset action
  and \val{tam.off.ent} without it; over the final \val{d.finalrows} logged updates, \val{tam.on.ent9} and
  \val{tam.off.ent9}. A uniform policy has \val{tam.ln7} nats (\DER).
\end{itemize}

\subsection{What the selected rewards encode}
\label{sec:genomes}

\begin{table}[t]
\centering
\small
\caption{\textbf{The frozen genomes.} Effective float32 parameters. ``Spent by one event'': a single occurrence
empties the need (relief at least $h_0=\val{d.h0}$) and the need never refills, so only the first such event in an
episode pays. A record of the frozen rewards; ``steps per full refill'' is $1/g$ (\DER). \src{table6}{results/table6_genomes.csv}}
\label{tab:genomes}
\input{tables/genomes.tex}
\end{table}

\begin{figure}[t]
\centering
\includegraphics[width=5.5in]{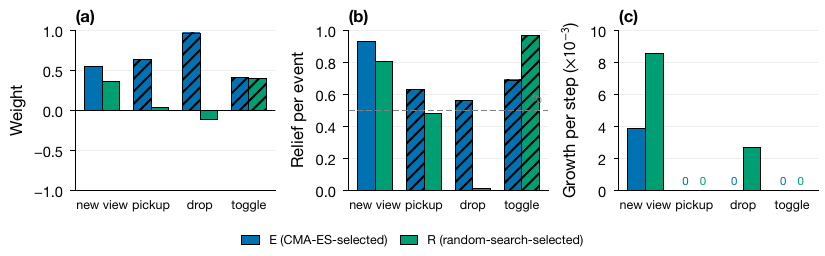}
\caption{\textbf{E pays its first pickup, drop and toggle once per episode; R slightly penalizes drops.} (a) Weights,
(b) reliefs (dashed: $h_0$) and (c) refill rates of the frozen E and R genomes, per event. Hatched bars mark events
whose single occurrence empties the need and whose need never refills. R's drop penalty is at most about
\val{g.R.droppenalty} of intrinsic reward per drop (\DER). A record of the frozen rewards. \src{figure7}{results/figure7_genomes.csv}}
\label{fig:genomes}
\end{figure}

\paragraph{The two selected genomes} (a record of the frozen rewards; Table~\ref{tab:genomes},
Figure~\ref{fig:genomes}).
\begin{itemize}[itemsep=1pt]
  \item E has positive weights on all four events. Only its view need refills (about every
  \val{g.E.newview.period} steps). One pickup, one drop and one toggle each empty their needs for the rest of the
  episode, so E pays the first of each.
  \item R has a small negative drop weight (\val{g.R.drop.w}) and a small drop relief (\val{g.R.drop.s}), so it
  slightly penalizes drops: a drop lowers its intrinsic reward by at most about \val{g.R.droppenalty} (\DER;
  $|w|\,s\,(2-s)$ for drop weight $w$ and relief $s$). Its drop and view needs refill. Its toggle need, like E's, is
  spent by one event; its pickup need takes two.
\end{itemize}

\paragraph{Genome features across all searched candidates} (\EXP; specified after the Stage~2 results were known; no
$p$-values; Table~\ref{tab:exploratory} in Appendix~\ref{app:exploratory}). Associations in an adaptively sampled
population are not effects of the features.
\begin{itemize}[itemsep=1pt]
  \item \textbf{Toggle and pickup sign.} A positive toggle weight went with higher mean combined fitness across all
  candidates (\val{x.pooled.toggle.positive} against \val{x.pooled.toggle.negative}) and among random-search candidates
  alone (\val{x.random.toggle.positive} against \val{x.random.toggle.negative}); pickup showed the same pattern.
  \item \textbf{Drop sign.} Across all candidates a positive drop weight went with higher fitness
  (\val{x.pooled.drop.positive} against \val{x.pooled.drop.negative}); among random-search candidates the order
  reversed (\val{x.random.drop.positive} against \val{x.random.drop.negative}). We treat the pooled drop association as
  unsupported.
  \item \textbf{Zero refill.} Zero refill on toggle and on pickup went with higher fitness in both populations, more
  weakly among random-search candidates; for drop the association reversed among random-search candidates.
\end{itemize}

\section{What our predictions and measures got wrong}
\label{sec:wrong}
Appendix~\ref{app:outcomes} lists every registered rule and test with its outcome. Of the \val{d.holm1} primary and
\val{d.holm2} secondary comparisons, \val{kc.hits} and \val{sec.hits} hit; the tamper prediction hit. The misses and
measurement failures were these.
\begin{itemize}[itemsep=2pt]
  \item \textbf{E over R on the training tasks: no registered difference.} The secondary family predicted that E
  would beat R on both training tasks. Both comparisons missed (Section~\ref{sec:training}); on DoorKey-8x8 R's mean
  area was the higher. On their re-evaluation panels (different seeds), R had also scored higher than E. Because E and R are single selected
  genomes, the misses say nothing general about CMA-ES and random search.
  \item \textbf{The tamper label's press threshold was too low.} The label asked for more than one press per
  completed episode; every final window cleared it many times over, while the per-step rate was below a uniform
  policy's, so a policy pressing at random would also have met it (\DER). The area half of the label holds; the press
  half measures little in long, failing episodes, as the earlier paper found for its own repeat threshold.
  \item \textbf{The search score overstated the CMA-ES finalists.} Not a registered prediction: the
  \val{s2.searchseeds}-seed search score ranked every CMA-ES finalist above every random-search finalist, and fresh seeds reversed that order
  (Table~\ref{tab:finalists}). The registered re-evaluation step caught it.
  \item \textbf{The pooled drop association} reversed among random-search candidates alone. One untested
  explanation (\HYP) is adaptive sampling: CMA-ES concentrated its later generations near its early winners, so their
  features are over-represented in the pooled population whether or not they matter.
  \item \textbf{Infrastructure.}
  \begin{itemize}[itemsep=1pt]
    \item A cross-platform resume of the Stage~2 controller failed exact replay: \val{s2.ulp} latent coordinates
    differed by one float64 unit in the last place between the Mac and Linux, and one decoded refill value by
    \val{s2.ulpgrowth}. No rounding or tolerance was introduced; the controller resumed on the original Mac after exact
    replay passed there (Appendix~\ref{app:infra}).
    \item Stage~2 orchestration depended on a laptop that had to stay awake and online. Two transport failures and a
    failed migration preceded the final restart; \val{s2.terminated} terminated calls were retried with identical
    inputs, and no saved result was recomputed.
    \item The Stage~3 check that the controller survives a closed laptop lid: we waived it; it remains unverified. The
    run completed without a controller restart.
  \end{itemize}
\end{itemize}

\section{Discussion}
\label{sec:discussion}

\paragraph{Searching over drives helped and transferred relative to the generic baselines.} E beat task-only PPO and
episodic novelty on both training tasks and on the held-out task, and beat R on the held-out task (\REG). The held-out
win over R compares two selected genomes; on the training tasks the two were not distinguishable by the registered
tests. Our reading (\EXP) is that the drives fit the structure of the training tasks: DoorKey-8x8 and UnlockPickup
each require a short chain of a key pickup and a door, and E pays the first pickup, drop and toggle of every episode.

\paragraph{Why S did better on KeyCorridorS3R2 (\HYP, untested).} KeyCorridorS3R2 has several closed doors. To
succeed, the agent must open the door of the side room that holds the key and then the locked door in front of the
target. E's toggle need is spent by the first door or box change of any kind, possibly a door the agent must open anyway,
and it never refills; R's toggle need behaves the same way. S's open event fires only when the initially locked door
opens, and its need keeps refilling until then. A plausible account is that E's spent-once door drive is used up before
the agent reaches the locked door, whereas S keeps being pulled toward it. This is untested: the saved per-episode
records do not include door events, and S differs from E and R on KeyCorridor in two confounded ways, hand-wired events
(key-only pickup, locked-door-only open) and a refilling door need. The UnlockPickup comparison, where S uses generic
events and did worse than E, is a different task and cannot separate the two.

\paragraph{Why random search sufficed (\EXP).} The candidate associations suggest a landscape dominated by a few
binary choices: a positive sign on the task events (pickup, toggle) and no refill on them. Under the random-search
prior each of these four choices has probability one half, so roughly \val{x.sixteen} of uniform draws gets all four
right (\DER; a heuristic that assumes these are the only choices that matter). With \val{s2.cands} draws that is
plenty, which is consistent with random search keeping pace with CMA-ES, as \citet{bergstra2012random} found for
hyperparameters where few dimensions matter. We did not test this.

\paragraph{Selection did not protect E against an exploit absent during selection.} The reset action was not part of
the search environment, so the search could neither reward nor penalize a reward's response to it, and E's positive
$P_0$ made a reset pay whenever the drive was positive. The registered collapse and the descriptive comparison with
task-only PPO are qualitatively consistent with the earlier paper's result for a hand-designed reward, where a signal the
agent could fill itself did worse than no signal. In the terms of \citet{everitt2021reward}, the
reset is tampering with the reward function's input, and nothing in our selection removed the incentive to do it.

\paragraph{Implications.} A selection environment can only select against failures it contains. Include the
exploits one anticipates, such as a reset, in the environments a reward is selected on, and select across
structurally different tasks (several doors, distractor objects) rather than two tasks that share one subgoal chain.

\section{Limitations}
\label{sec:limitations}
\begin{itemize}[itemsep=1pt]
  \item \textbf{Tasks.} One held-out task and \val{d.ntrain} training tasks, all in MiniGrid. The transfer claim rests
  on KeyCorridorS3R2 alone.
  \item \textbf{Search noise and learner.} \val{s2.searchseeds} seeds per task per candidate, with generation-specific
  seed panels; one PPO configuration, the earlier paper's (Appendix~\ref{app:hyper}).
  \item \textbf{Numerics and infrastructure.} Same-seed runs are not bit-identical across CPU types, so comparisons are
  unpaired and exact cross-machine reproduction is not established (Appendix~\ref{app:infra}); Stage~2 had
  interruptions, retries and a laptop-bound controller.
  \item \textbf{Comparators.} S is descriptive only and is confounded on KeyCorridor (hand-wired events and a refilling
  door need); on UnlockPickup, with generic events, it scored below E. E against R compares two selected genomes, not two search methods.
  \item \textbf{Exploratory analyses} are post hoc, in an adaptively sampled population.
  \item \textbf{Tamper scope.} DoorKey-8x8 only, free presses, E only.
  \item \textbf{Detector.} Event definitions differ from the earlier paper's archived detector; comparisons with the
  earlier paper are qualitative.
\end{itemize}

\section{Conclusion}
\label{sec:conclusion}
A search over four generic drives found a reward that helped PPO on its training tasks and on one held-out task,
where it beat task-only PPO, episodic novelty and the random-search selection. A satiation reward told which events
matter on that task did better still, although E outscored it on UnlockPickup, where it used generic events. The two selected rewards were not distinguishable on the tasks they were
selected on. And the searched reward gave no protection against a reset action that the search never saw: learning
collapsed, below that of task-only PPO. Searching over drives can find useful ones, but it selects for the
environments it is shown, including what they leave out.

\phantomsection\label{sec:references}
\bibliographystyle{plainnat}
\bibliography{references}

\clearpage
\appendix

\section{Every registered rule and test, Stages 1--3}
\label{app:outcomes}

\begin{table}[h]
\centering
\footnotesize
\caption{\textbf{Registered rules and their outcomes.} Stage~1 rules come from the locked Stage~1 protocol; Stage~2
and Stage~3 rules from the locked Stage~2 protocol. The Stage~1 separation test was registered as a two-sided,
descriptive test with no prediction. \src{table7}{results/table7_outcomes.csv}}
\label{tab:outcomes}
\input{tables/outcomes.tex}
\end{table}

The Stage~1 separation $p$-values (two-sided, Holm-adjusted over three tasks) were \val{s1.sep.DoorKey8x8.ph} for
DoorKey-8x8, \val{s1.sep.KeyCorridorS3R1.ph} for KeyCorridorS3R1 and \val{s1.sep.UnlockPickup.ph} for UnlockPickup.

\section{Amendment and approval timeline}
\label{app:timeline}

\begin{table}[h]
\centering
\scriptsize
\setlength{\tabcolsep}{3pt}
\caption{\textbf{Every amendment.} Creation times are recorded inside each file. ``First public push'' is the earliest
push of a commit containing the amendment, from GitHub's public event feed (feed) or, where the feed did not reach,
from the public-commit receipts verified before the Stage~3 launch (receipt). The last column prints each file's
\texttt{scientific\_changes} field. \src{table8}{results/table8_amendments.csv}}
\label{tab:amendments}
\input{tables/amendments.tex}
\end{table}

\begin{itemize}[itemsep=1pt]
  \item The Stage~2 lock (\val{reg.s2.lock}, commit \val{reg.s2.commit}, tested commit \val{reg.s2.tested}) was pushed
  \val{reg.s2.push}; the public ref was observed at \val{reg.s2.obs}.
  \item The budget-\val{cost.s2.cap} and Mac-resume amendments are infrastructure and budget changes, not scientific
  ones (their \texttt{scientific\_changes} fields are false). The gate report recorded, without a time, that they were
  pushed before the restart checks, and the restarted controller's start-up was verified at \val{reg.macrestart}.
  GitHub's public event feed lists their commit's push in Table~\ref{tab:amendments}.
  \item The Stage~3 descriptive amendment (SHA256 prefix \val{reg.descsha}) and the exploratory plan were committed
  together; the descriptive amendment was bound by both public receipts (\val{reg.r1.verified} and
  \val{reg.r2.verified}) before the science run started (\val{reg.launch}). The exploratory plan was written after
  the Stage~2 results were known, so it pre-registers nothing about them.
\end{itemize}

\section{Infrastructure history}
\label{app:infra}
\begin{itemize}[itemsep=1pt]
  \item \textbf{Stage~2.} \val{s2.calls} provider calls and \val{s2.attempts} physical attempts; \val{s2.extra}
  additional attempts on existing calls (restarts, cause not established) and \val{s2.manual} same-input manual retries
  of the \val{s2.terminated} terminated calls; usage was missing for \val{s2.nousage} attempts. \val{s2.ctrl}
  controller events: two transport failures (a download name-resolution failure and an output-polling deadline), a
  failed Linux replay, a read-only Linux diagnosis and the final Mac restart. Successful outputs were recovered, not
  recomputed. The Linux replay failed because latent coordinates differed by one unit in the last place; the
  diagnosis confirmed numeric differences rather than metadata.
  \item \textbf{Stage~3.} \val{s3.handles} provider calls, of which \val{s3.runs} were science runs and the rest pre-launch
  smoke checks; \val{s3.extra} additional same-call attempts (provider
  restarts under the same handle), \val{s3.manual} manual retries and \val{s3.ctrl} controller restarts. An integrity
  check verified all \val{s3.files} result files listed in the completion markers and the \val{s3.srcfiles}
  registered source files.
  \item \textbf{CPU types.} Successful Stage~2 scientific attempts ran on Zen~3 (\val{cpu.s2.zen3}), Intel
  (\val{cpu.s2.intel}), Zen~4 (\val{cpu.s2.zen4}) and other (\val{cpu.s2.other}) CPUs; Stage~3 science attempts on
  Zen~3 (\val{cpu.s3.zen3}), Intel (\val{cpu.s3.intel}), Zen~4 (\val{cpu.s3.zen4}) and other (\val{cpu.s3.other}).
  Stage~1 found the same seed \val{s1.det} across containers. A pinned pre-check (\val{pin.n} replicates: Zen~3
  \val{pin.zen3}, Zen~4 \val{pin.zen4}, Intel \val{pin.intel}) produced \val{pin.fp} distinct fingerprints, one shared
  by the AMD containers and one by the Intel containers; the math-library pin was honoured only on Intel. Runs were
  therefore left unpinned and compared unpaired.
\end{itemize}

\section{Cost accounting}
\label{app:cost}

\begin{table}[h]
\centering
\small
\caption{\textbf{Spending by stage, in three quantities that overlap and must not be added.} The guard is a
conservative reservation total, not a charge. The measured compute estimate multiplies requested resources by measured
wall time. Provider billing covers complete billing intervals only and is not a final invoice. \src{table9}{results/table9_costs.csv}}
\label{tab:costs}
\input{tables/costs.tex}
\end{table}

\section{Exploratory tables}
\label{app:exploratory}

\begin{table}[h]
\centering
\footnotesize
\caption{\textbf{Genome features and combined search fitness (\EXP).} Mean combined fitness by weight sign and by
zero versus nonzero refill, for all \val{s2.candtotal} candidates and for the \val{s2.cands} random-search candidates
alone, with group sizes; Spearman correlation of weight with combined fitness. No $p$-values were computed. Full
tables, including per-task outcomes and every feature, are in the Stage~3 report archive. \src{table10}{results/table10_exploratory.csv}}
\label{tab:exploratory}
\input{tables/exploratory.tex}
\end{table}

\section{Hyperparameters, software and the episode ceiling}
\label{app:hyper}

\begin{table}[h]
\centering
\small
\caption{\textbf{PPO settings (the earlier paper's) and software versions.} From the archived trainer configuration and
the locked environment. \src{table11}{results/table11_hyper.csv}}
\label{tab:hyper}
\input{tables/hyper.tex}
\end{table}

\paragraph{The episode ceiling.} Write $V_t=V(h_t)$. Summing $r^\text{int}_t=V_t-V_{t+1}$ with discount $\gamma\in(0,1]$
over an episode of length $T$ gives
$G=\sum_{t=0}^{T-1}\gamma^t(V_t-V_{t+1})=V_0-(1-\gamma)\sum_{t=1}^{T-1}\gamma^{t-1}V_t-\gamma^{T-1}V_T$.
The coefficients on $V_1,\dots,V_T$ are non-negative and sum to one, so $G$ lies between $V_0-\max_hV(h)$ and
$V_0-\min_hV(h)$. Over $h\in[0,1]^4$, $\max_hV(h)=\sum_k\max(w_k,0)$ and $\min_hV(h)=\sum_k\min(w_k,0)$. A reset
sets $h$ to zero, a valid need state, so the bound holds with the reset action as well. For E every weight is positive,
so the upper bound is $V_0=P_0$.

\section{Reproducibility}
\label{app:repro}
\begin{itemize}[itemsep=1pt]
  \item \textbf{Repository.} \url{https://github.com/chasemetoyer/found-by-search}, tag
  \texttt{v1.1}, archived on Zenodo at \href{https://doi.org/10.5281/zenodo.23828842}{doi:10.5281/zenodo.23828842}. \nolinkurl{README.md} maps every table, figure and headline number to its \texttt{make}
  target and its committed file in \nolinkurl{results/}; \texttt{make all} rebuilds the numbers, tables, figures and
  this PDF from the committed data in seconds on a laptop, and \texttt{make check} verifies them.
  \item \textbf{Numbers.} Every value is a macro generated by \nolinkurl{scripts/make_numbers.py} from snapshots of the
  saved result files in \nolinkurl{data/raw} (local paths replaced by placeholders; SHA256 sums in
  \nolinkurl{data/raw/SHA256SUMS}) and links to its line in \nolinkurl{results/numbers.json}.
  \nolinkurl{scripts/check_numbers.py} regenerates every macro and table, cross-checks headline values against an
  independent extraction from the raw run logs (\nolinkurl{data/extracted/independent-extraction.json}), and rejects
  literal numbers in the prose.
  \item \textbf{Training code and registration.} The locked protocols, every amendment, the public-commit receipts,
  the frozen winners and the exact training code are in \href{\repofile{research}}{\nolinkurl{research/}}, copied from
  commit \val{reg.research} of the project's working repository (\texttt{make verify-lock} checks every locked file
  against its registered hash); the files the numbers cite are also copied into
  \nolinkurl{data/raw/repo}. The Stage~2 lock is commit \val{reg.s2.commit}; the Stage~3 science run used source
  commit \val{s3.commit}. The full Stage~3 run archive (run logs and policy checkpoints) is not in the repository.
  \item \textbf{Archive.} The repository at tag \texttt{v1.1} is archived on Zenodo at
  \href{https://doi.org/10.5281/zenodo.23828842}{doi:10.5281/zenodo.23828842}. The earlier paper's release materials are at
  doi:10.5281/zenodo.23708846.
\end{itemize}

\end{document}

%% file: tables/calibration.tex
\begin{tabular}{@{}lllrll@{}}
\toprule
Task & Candidate role & Budget & Best baseline area & Outcome & Reason \\
\midrule
DoorKey-8x8 & training & 1M & 0.195 & kept & -- \\
KeyCorridorS3R1 & training & 1M & 0.855 & rejected & too easy (area $>0.85$) \\
UnlockPickup & training & 1M & 0.006 & kept & -- \\
KeyCorridorS3R2 & held out & 2M & 0.053 & kept & -- \\
BlockedUnlockPickup & held out & 2M & 0.008 & rejected & too hard (area $<0.05$) \\
DoorKey-16x16 & held out & 2M & 0.045 & rejected & too hard (area $<0.05$) \\
MultiRoom-N2-S4 & held out & 2M & 0.961 & rejected & too easy (area $>0.85$) \\
\bottomrule
\end{tabular}

%% file: tables/finalists.tex
\begin{tabular}{@{}llrrrrrrr@{}}
\toprule
Method & Candidate & Search & Re-eval. & Change & DK area & DK solved & UP area & UP solved \\
\midrule
CMA-ES & 92 & 0.707 & 0.587 & $-$0.120 & 0.839 & 10/10 & 0.336 & 5/10 \\
CMA-ES & 113 (\textbf{E}) & 0.692 & 0.596 & $-$0.096 & 0.818 & 10/10 & 0.373 & 3/10 \\
CMA-ES & 94 & 0.676 & 0.497 & $-$0.179 & 0.766 & 9/10 & 0.227 & 2/10 \\
Random search & 57 (\textbf{R}) & 0.610 & 0.632 & +0.022 & 0.875 & 10/10 & 0.389 & 4/10 \\
Random search & 116 & 0.584 & 0.626 & +0.042 & 0.911 & 10/10 & 0.340 & 4/10 \\
Random search & 103 & 0.510 & 0.619 & +0.108 & 0.870 & 10/10 & 0.368 & 5/10 \\
\bottomrule
\end{tabular}

%% file: tables/final.tex
\begin{tabular}{@{}lrrrr@{}}
\toprule
Method & Mean area (SD) & Solved & 95\% CI (solved) & Median first solve \\
\midrule
\multicolumn{5}{@{}l}{\textit{KeyCorridorS3R2 (held out)}} \\
E (CMA-ES-selected) & 0.457 (0.139) & 4/20 & [0.06, 0.44] & 2M (censored) \\
R (random-search-selected) & 0.298 (0.102) & 0/20 & [0.00, 0.17] & 2M (censored) \\
T (task-only) & 0.004 (0.002) & 0/20 & [0.00, 0.17] & 2M (censored) \\
N (episodic novelty) & 0.064 (0.035) & 0/20 & [0.00, 0.17] & 2M (censored) \\
S (satiation)$^\dagger$ & 0.768 (0.046) & 20/20 & [0.83, 1.00] & 0.63M \\
\addlinespace
\multicolumn{5}{@{}l}{\textit{DoorKey-8x8 (training)}} \\
E (CMA-ES-selected) & 0.837 (0.058) & 20/20 & [0.83, 1.00] & 0.45M \\
R (random-search-selected) & 0.867 (0.048) & 20/20 & [0.83, 1.00] & 0.37M \\
T (task-only) & 0.286 (0.333) & 8/20 & [0.19, 0.64] & 2M (censored) \\
N (episodic novelty) & 0.603 (0.226) & 17/20 & [0.62, 0.97] & 0.86M \\
S (satiation)$^\dagger$ & 0.903 (0.031) & 20/20 & [0.83, 1.00] & 0.31M \\
\addlinespace
\multicolumn{5}{@{}l}{\textit{UnlockPickup (training)}} \\
E (CMA-ES-selected) & 0.349 (0.187) & 8/20 & [0.19, 0.64] & 1.68M \\
R (random-search-selected) & 0.294 (0.134) & 3/20 & [0.03, 0.38] & 2M (censored) \\
T (task-only) & 0.003 (0.002) & 0/20 & [0.00, 0.17] & 2M (censored) \\
N (episodic novelty) & 0.006 (0.002) & 0/20 & [0.00, 0.17] & 2M (censored) \\
S (satiation)$^\dagger$ & 0.073 (0.143) & 1/20 & [0.00, 0.25] & 2M (censored) \\
\bottomrule
\end{tabular}

%% file: tables/comparisons.tex
\begin{tabular}{@{}llrrrrllr@{}}
\toprule
Task & Test & Mean gap & $U$ & Raw $p$ & Holm $p$ & Rank test & Meaningful & Fisher $p$ \\
\midrule
\multicolumn{9}{@{}l}{\textit{Primary family, held-out task (Holm, $m$=3)}} \\
KeyCorridorS3R2 & E $>$ T & +0.452 & 400.0 & $5\times10^{-6}$ & $1\times10^{-5}$ & \textbf{hit} & yes & 0.053 \\
KeyCorridorS3R2 & E $>$ N & +0.393 & 400.0 & $7.3\times10^{-12}$ & $2.2\times10^{-11}$ & \textbf{hit} & yes & 0.053 \\
KeyCorridorS3R2 & E $>$ R & +0.159 & 329.0 & 0.00015 & 0.00015 & \textbf{hit} & yes & 0.053 \\
\addlinespace
\multicolumn{9}{@{}l}{\textit{Secondary family, training tasks (Holm, $m$=6)}} \\
DoorKey-8x8 & E $>$ T & +0.551 & 389.0 & $1.4\times10^{-9}$ & $8.5\times10^{-9}$ & \textbf{hit} & yes & $2.3\times10^{-5}$ \\
DoorKey-8x8 & E $>$ N & +0.234 & 358.0 & $2.1\times10^{-6}$ & $1\times10^{-5}$ & \textbf{hit} & yes & 0.115 \\
DoorKey-8x8 & E $>$ R & $-$0.031 & 124.0 & 0.981 & 0.981 & \textbf{miss} & no & 1.000 \\
UnlockPickup & E $>$ T & +0.346 & 399.0 & $5\times10^{-6}$ & $2\times10^{-5}$ & \textbf{hit} & yes & 0.002 \\
UnlockPickup & E $>$ N & +0.342 & 396.0 & $5\times10^{-6}$ & $2\times10^{-5}$ & \textbf{hit} & yes & 0.002 \\
UnlockPickup & E $>$ R & +0.055 & 233.0 & 0.192 & 0.383 & \textbf{miss} & no & 0.078 \\
\bottomrule
\end{tabular}

%% file: tables/tamper.tex
\begin{tabular}{@{}llrr@{}}
\toprule
Endpoint & Level & Tamper off & Tamper on \\
\midrule
Signed start payoff $P_0$ (fixed before data) & REG & \multicolumn{2}{c}{0.6396 $>0$: collapse branch} \\
Mean curve area & REG & 0.829 & 0.136 \\
Off minus on; one-sided $U$, $p$ & REG & \multicolumn{2}{c}{0.693; $U$=393.0, $p$=$5\times10^{-6}$} \\
Solved seeds & PSD & 20/20 & 4/20 \\
Presses per step, all training & PSD & 1.4\% & 10.3\% \\
Presses per step, final 10\% & PSD & 0.004\% & 7.9\% \\
Presses per completed episode, all training & PSD & 0.6 & 52.4 \\
Presses per completed episode, final 10\% & PSD & 0.0 & 39.4 \\
Mean episode intrinsic return & PSD & 0.592 & 0.637 \\
Largest episode intrinsic return & PSD & 0.63964 & 0.63964 \\
Episodes above exact $P_0$ & PSD & 1 & 0 \\
Policy entropy, all updates (nats) & PSD & 0.787 & 1.634 \\
Policy entropy, final 9 updates (nats) & PSD & 0.387 & 1.516 \\
\bottomrule
\end{tabular}

%% file: tables/genomes.tex
\begin{tabular}{@{}llrrrrl@{}}
\toprule
Reward & Event & Weight & Relief & Growth per step & Steps per full refill & Spent by one event \\
\midrule
E & new\_view & 0.554 & 0.928 & 0.0039 & 258 & no \\
 & pickup & 0.632 & 0.633 & 0 & -- & yes \\
 & drop & 0.966 & 0.564 & 0 & -- & yes \\
 & toggle & 0.406 & 0.690 & 0 & -- & yes \\
\addlinespace
R & new\_view & 0.368 & 0.803 & 0.0086 & 117 & no \\
 & pickup & 0.039 & 0.483 & 0 & -- & no \\
 & drop & $-$0.106 & 0.014 & 0.0027 & 367 & no \\
 & toggle & 0.399 & 0.966 & 0 & -- & yes \\
\bottomrule
\end{tabular}

%% file: tables/outcomes.tex
\begin{tabular}{@{}p{5.4cm}p{6.4cm}p{3.6cm}@{}}
\toprule
Registered rule or test & Result & Outcome \\
\midrule
\multicolumn{3}{@{}l}{\textit{Stage 1 (locked 28 October 2026, 03:12:23~UTC)}} \\
Validation gate before training & 53 tests, 0 failures & passed \\
Determinism check (report, no tuning) & same seed, separate containers & not bit-identical \\
Task-selection rule & kept: DoorKey-8x8, UnlockPickup, KeyCorridorS3R2 & applied mechanically \\
 & rejected: KeyCorridorS3R1, BlockedUnlockPickup, DoorKey-16x16, MultiRoom-N2-S4 &  \\
1M baseline separation (two-sided, Holm $m$=3) & descriptive test, no prediction & not separated in 3 of 3 \\
Spending limit & \$15; guard \$2.88 & within limit \\
\midrule
\multicolumn{3}{@{}l}{\textit{Stage 2 (locked 28 October 2026, 14:28:30~UTC)}} \\
Equal candidate budget & 120 CMA-ES, 120 random & held \\
Generation barrier and exact ask/tell replay & replay reproduces every CMA-ES candidate & held \\
Candidate-failure guard ($>$5\% stops search) & 0 candidate numerical failures & not triggered \\
Winner rule (highest re-evaluated mean area) & E = candidate 113, R = candidate 57 & applied \\
Budget stop rule & cap \$35 raised to \$50 by amendment; guard \$29.41 & within cap \\
\midrule
\multicolumn{3}{@{}l}{\textit{Stage 3 (registered in the Stage 2 lock)}} \\
primary: E $>$ T, KeyCorridorS3R2 & Holm $p$ = $1\times10^{-5}$; gap +0.452 & hit; meaningful \\
primary: E $>$ N, KeyCorridorS3R2 & Holm $p$ = $2.2\times10^{-11}$; gap +0.393 & hit; meaningful \\
primary: E $>$ R, KeyCorridorS3R2 & Holm $p$ = 0.00015; gap +0.159 & hit; meaningful \\
secondary: E $>$ T, DoorKey-8x8 & Holm $p$ = $8.5\times10^{-9}$; gap +0.551 & hit; meaningful \\
secondary: E $>$ N, DoorKey-8x8 & Holm $p$ = $1\times10^{-5}$; gap +0.234 & hit; meaningful \\
secondary: E $>$ R, DoorKey-8x8 & Holm $p$ = 0.981; gap $-$0.031 & \textbf{miss} \\
secondary: E $>$ T, UnlockPickup & Holm $p$ = $2\times10^{-5}$; gap +0.346 & hit; meaningful \\
secondary: E $>$ N, UnlockPickup & Holm $p$ = $2\times10^{-5}$; gap +0.342 & hit; meaningful \\
secondary: E $>$ R, UnlockPickup & Holm $p$ = 0.383; gap +0.055 & \textbf{miss} \\
Tamper branch fixed by sign of $P_0$ & $P_0$ = 0.6396 & collapse branch \\
Collapse test (off $>$ on) & $p$ = $5\times10^{-6}$ & hit \\
Budget stop rule & cap \$10 raised to \$15 by amendment; guard \$7.04 & within cap \\
\bottomrule
\end{tabular}

%% file: tables/amendments.tex
\begin{tabular}{@{}lllll@{}}
\toprule
Amendment and scope & Created (UTC) & Commit & First public push (UTC) & Scientific change \\
\midrule
S1: infrastructure & 28 Oct, 03:20:36 & 8f22e05 & 03:20:38 (feed) & false \\
S1: infrastructure & 28 Oct, 03:24:33 & b34b2c4 & 03:24:34 (feed) & false \\
S1: setup repairs & 28 Oct, 03:30:52 & 196c525 & 03:30:53 (feed) & false \\
S1: report only & 28 Oct, 03:45:10 & 373ce9f & 04:23:26 (feed) & false \\
S1: report only & 28 Oct, 04:36:22 & 69f50f3 & 04:36:45 (feed) & false \\
S2: read-only dashboard & 28 Oct, 16:19:47 & b48e68b & 16:19:50 (feed) & false \\
S2: transport recovery & 28 Oct, 19:40:25 & 739b3ad & 19:45:13 (feed) & false \\
S2: polling recovery & 28 Oct, 20:39:30 & c5541f8 & 20:43:21 (feed) & false \\
S2: cloud controller & 28 Oct, 21:17:33 & d235a13 & 21:23:17 (feed) & false \\
S2: budget cap to \$50 & 28 Oct, 22:11:22 & 3cfaa88 & 22:15:56 (feed) & false \\
S2: Mac controller resume & 28 Oct, 22:11:22 & 3cfaa88 & 22:15:56 (feed) & false \\
S3: budget cap to \$15 & 29 Oct, 14:25:26 & 7b15f3a & 15:06:14 (receipt) & false \\
S3: pre-launch update & 29 Oct, 14:29:56 & 5c5c406 & 15:06:14 (receipt) & false \\
S3: pre-launch hardening & 29 Oct, 15:06:12 & 490cf87 & 15:06:14 (receipt) & false \\
S3: infrastructure (guardian name) & 29 Oct, 15:46:18 & 44198cf & 15:57:35 (receipt) & false \\
S3: infrastructure (web concurrency) & 29 Oct, 15:57:29 & cd79ce5 & 15:57:35 (receipt) & false \\
S3: descriptive endpoints & 29 Oct, 14:27:03 (commit) & 7b15f3a & 15:06:14 (receipt) & n/a \\
S3: exploratory plan (after Stage 2 results) & 29 Oct, 14:27:03 (commit) & 7b15f3a & 15:06:14 (receipt) & n/a \\
\bottomrule
\end{tabular}

%% file: tables/costs.tex
\begin{tabular}{@{}lrrrr@{}}
\toprule
Stage & Cap & Guard & Measured compute & Provider billing \\
\midrule
Stage 1 & \$15 & \$2.88 & -- & \$1.52 \\
Stage 2 & \$35 $\to$ \$50 & \$29.41 & \$22.37 & \$22.07 \\
Stage 3 & \$10 $\to$ \$15 & \$7.04 & \$4.81 & \$5.08 \\
\bottomrule
\end{tabular}

%% file: tables/exploratory.tex
\begin{tabular}{@{}lllll@{}}
\toprule
Event & Feature & Groups & All candidates: means (n) & Random search: means (n) \\
\midrule
toggle & weight sign & positive vs negative & 0.351 vs 0.075 (160/80) & 0.204 vs 0.068 (61/59) \\
 & refill & zero vs nonzero & 0.307 vs 0.170 (156/84) & 0.160 vs 0.110 (66/54) \\
 & Spearman $\rho$, weight &  & 0.53 & 0.21 \\
\addlinespace
pickup & weight sign & positive vs negative & 0.320 vs 0.106 (172/68) & 0.183 vs 0.071 (71/49) \\
 & refill & zero vs nonzero & 0.338 vs 0.156 (136/104) & 0.176 vs 0.107 (53/67) \\
 & Spearman $\rho$, weight &  & 0.37 & 0.33 \\
\addlinespace
drop & weight sign & positive vs negative & 0.296 vs 0.201 (147/93) & 0.105 vs 0.171 (61/59) \\
 & refill & zero vs nonzero & 0.299 vs 0.197 (146/94) & 0.133 vs 0.142 (59/61) \\
 & Spearman $\rho$, weight &  & 0.27 & $-$0.13 \\
\bottomrule
\end{tabular}

%% file: tables/hyper.tex
\begin{tabular}{@{}ll@{}}
\toprule
Setting & Value \\
\midrule
Parallel environments & 16 \\
Rollout length per environment & 128 \\
Learning rate (linearly annealed) & 0.00025 \\
Discount $\gamma$ & 0.99 \\
GAE $\lambda$ & 0.95 \\
Minibatches per update & 4 \\
Epochs per update & 4 \\
Clip coefficient & 0.2 \\
Entropy coefficient & 0.01 \\
Value coefficient & 0.5 \\
Max gradient norm & 0.5 \\
Advantage normalization & per minibatch \\
Evaluation episodes per run & 64 \\
\midrule
Python & 3.12.12 \\
torch & 2.14.1+cpu \\
minigrid & 3.2.0 \\
gymnasium & 1.4.0 \\
numpy & 2.5.4 \\
cma & 4.5.0 \\
scipy & 1.17.1 \\
modal & 1.6.3 \\
\bottomrule
\end{tabular}